\documentclass[journal,twoside,web]{ieeecolor}
\usepackage{generic}
\usepackage{cite}
\usepackage{amsmath,amssymb,amsfonts}
\usepackage{algorithmic}
\usepackage{graphicx}
\usepackage{textcomp}
\usepackage[utf8]{inputenc} % allow utf-8 input
\usepackage[T1]{fontenc}    % use 8-bit T1 fonts
\usepackage{url}            % simple URL typesetting
\usepackage{booktabs}       % professional-quality tables
\usepackage{amsfonts}       % blackboard math symbols
\usepackage{nicefrac}       % compact symbols for 1/2, etc.
\usepackage{microtype}      % microtypography
\usepackage{multirow}
\usepackage{colortbl}
\usepackage[table,xcdraw]{xcolor}
\usepackage{makecell}
\usepackage{graphicx}
\usepackage{amsmath}
\usepackage{amssymb}
\usepackage{booktabs}
\usepackage{bm}
\usepackage[ruled,vlined]{algorithm2e}
\usepackage{wasysym}
\usepackage{dsfont}
\usepackage{hyperref}

\def\BibTeX{{\rm B\kern-.05em{\sc i\kern-.025em b}\kern-.08em
    T\kern-.1667em\lower.7ex\hbox{E}\kern-.125emX}}
\author{Shiao Xie, Hongyi Wang, Ziwei Niu, Hao Sun, Shuyi Ouyang, Yen-Wei Chen, \IEEEmembership{Member, IEEE}, and Lanfen Lin, \IEEEmembership{Member, IEEE}
\thanks{Shiao Xie and Hongyi Wang contributed equally to this work.}
\thanks{Shiao Xie, Hongyi Wang, Ziwei Niu are with the College of Computer Science and Technology, Zhejiang University, Hangzhou 310063, China (e-mail: 22160144@zju.edu.cn, whongyi@zju.edu.cn, nzw@zju.edu.cn).}
\thanks{Hao Sun, Shuyi Ouyang are with the College of Computer Science and Technology, Zhejiang University, Hangzhou 310063, China (e-mail: sunhaoxx@zju.edu.cn, oysy@zju.edu.cn).}
\thanks{Yen-Wei Chen is with the College of Information Science and Engineering, Ritsumeikan University, Kusatsu 5250058, Japan, and also with the College of Computer Science and Technology, Zhejiang University, Hangzhou 310063, China (e-mail: chen@is.ritsumei.ac.jp).}
\thanks{Lanfen Lin is with the College of Computer Science and Technology, Zhejiang University, Hangzhou 310063, China (e-mail: llf@zju.edu.cn).}}
\begin{document}
\title{SemSim: Revisiting Weak-to-Strong Consistency from a Semantic Similarity Perspective for Semi-supervised Medical Image Segmentation}

\maketitle
\begin{abstract}
Semi-supervised learning (SSL) for medical image segmentation is a challenging yet highly practical task, which reduces reliance on large-scale labeled dataset by leveraging unlabeled samples. 
Among SSL techniques, the weak-to-strong consistency framework, popularized by FixMatch, has emerged as a state-of-the-art method in classification tasks. 
Notably, such a simple pipeline has also shown competitive performance in medical image segmentation. 
However, two key limitations still persist, impeding its efficient adaptation:
(1) the neglect of contextual dependencies results in inconsistent predictions for similar semantic features, leading to incomplete object segmentation;
(2) the lack of exploitation of semantic similarity between labeled and unlabeled data induces considerable class-distribution discrepancy. 
To address these limitations, we propose a novel semi-supervised framework based on FixMatch, named SemSim, powered by two appealing designs from semantic similarity perspective: 
(1) rectifying pixel-wise prediction by reasoning about the intra-image pair-wise affinity map, thus integrating contextual dependencies explicitly into the final prediction; 
(2) bridging labeled and unlabeled data via a feature querying mechanism for compact class representation learning, which fully considers cross-image anatomical similarities.
As the reliable semantic similarity extraction depends on robust features, we further introduce an effective spatial-aware fusion module (SFM) to explore distinctive information from multiple scales. Extensive experiments show that SemSim yields consistent improvements over the state-of-the-art methods across three public segmentation benchmarks.
\end{abstract}


\section{Introduction}
\label{sec:introduction}
Medical image segmentation plays a critical role in various clinical scenarios, such as disease diagnosis and preoperative assessment. In recent years, Convolutional Neural Networks (CNNs) have advanced the progress of this dense prediction task through the development of various segmentation models~\cite{fcn,unet, unet++, deeplabv3+}. However, the laborious and time-consuming nature of manual annotation often results in a shortage of labeled data, impeding further progress in performance enhancement. Semi-supervised learning (SSL) has emerged as a notable solution, as it enables the utilization of large volume of unlabeled data, thus reducing the annotation burden significantly.
\begin{figure}[t]
\centering
\includegraphics[width=1\linewidth]{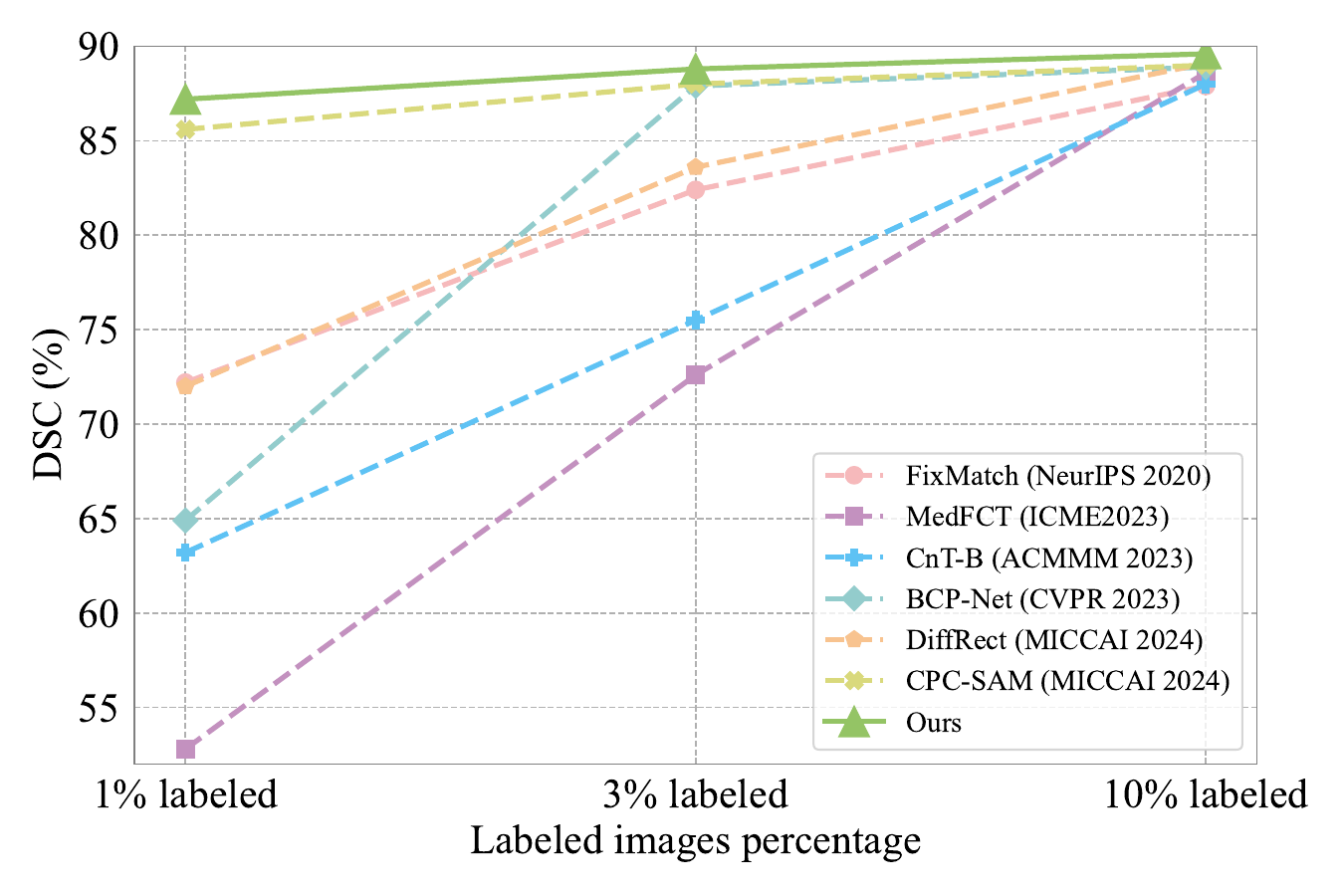}
\vspace{-20pt}
\caption{Comparison of state-of-the-art methods with SemSim on the ACDC dataset under different labeled ratios.}
\label{fig1}
\vspace{-10pt}
\end{figure}

\begin{figure}[t]
\centering
\includegraphics[width=1.0\linewidth, height=0.13\textheight]{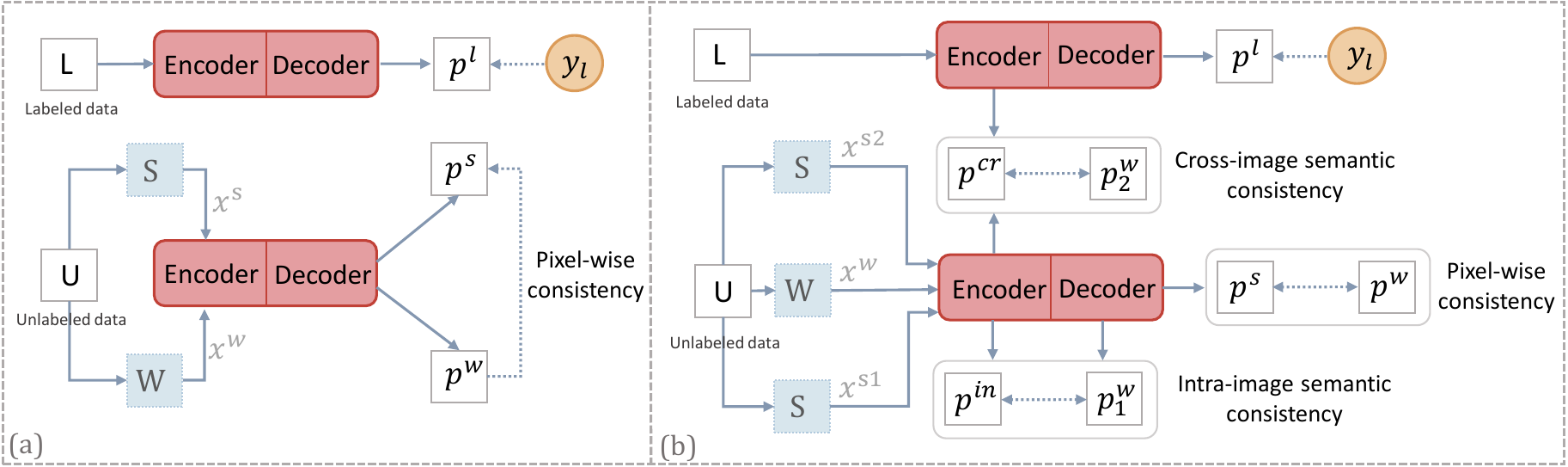}
\setlength{\abovecaptionskip}{4pt}
\vspace{-10pt}
\caption{Comparison of (a) FixMatch with (b) SemSim. $S$ and $W$ are strong and weak augmentations. $y_l$ represents the label and $p^l$ is the prediction of the labeled data.}
\label{fig2}
\vspace{-15pt}
\end{figure}

Several techniques are commonly used in SSL, including entropy minimization~\cite{EM}, pseudo-labeling~\cite{CPS, CTCT,s4cvnet}, and consistency regularization~\cite{MT, UAMT, ICT, ict_med, URPC, sohn2020fixmatch, SSNet}. In this realm, FixMatch~\cite{sohn2020fixmatch} has garnered significant attention in classification tasks, which is grounded in weak-to-strong consistency regularization. Interestingly, the experimental results in Fig.~\ref{fig1} reveal that it also achieves comparable performance on medical image segmentation benchmarks, particularly in scenarios with extremely limited labeled data. Thus, we select this simple yet effective framework as our baseline. As shown in Fig.~\ref{fig2} (a), when adapting FixMatch~\cite{sohn2020fixmatch} to segmentation tasks, the prediction $p^w$ of the weak augmented view $x^w$ acts as pseudo-label, enforcing pixel-wise consistency on the prediction $p^s$ of the strong augmented view $x^s$.  Intuitively, the success of this framework lies in generating high-quality $p^w$.

% Comparison of (a) MT-based SSL with (b) SemiCVT
% 强调下1%情况下fixmatch甚至是sota
% (a) The FixMatch baseline~\cite{sohn2020fixmatch}. (b) The proposed SemSim. $S$ and $W$ are strong and weak augmentation in the image-level. $y_l$ represents the label and $p^l$ represents the prediction for the labeled data. They are constrained through supervised loss.
% We establish a connection between labeled and unlabeled data by leveraging cross-image semantic feature similarity, generating the prediction $p^{cr}$. 
% As shown in Fig.~\ref{fig3}, through the weak-to-strong consistency constraint, such process in turn fully exploits reliable knowledge from labeled data thereby promoting a more compact intra-class distribution. 
% Given that the feature-level affinity map build contextual dependencies, we propose extracting these relations to refine pixel-wise predictions, resulting in $p^{in}$. Applying the constraint to $p^{in}$ provides explicit guidance for deep semantic features, thus enhancing the feature continuity in Fig.~\ref{fig3}. 
% The similarity of anatomical structures

\begin{figure}[t]
\centering
\includegraphics[width=1\linewidth]{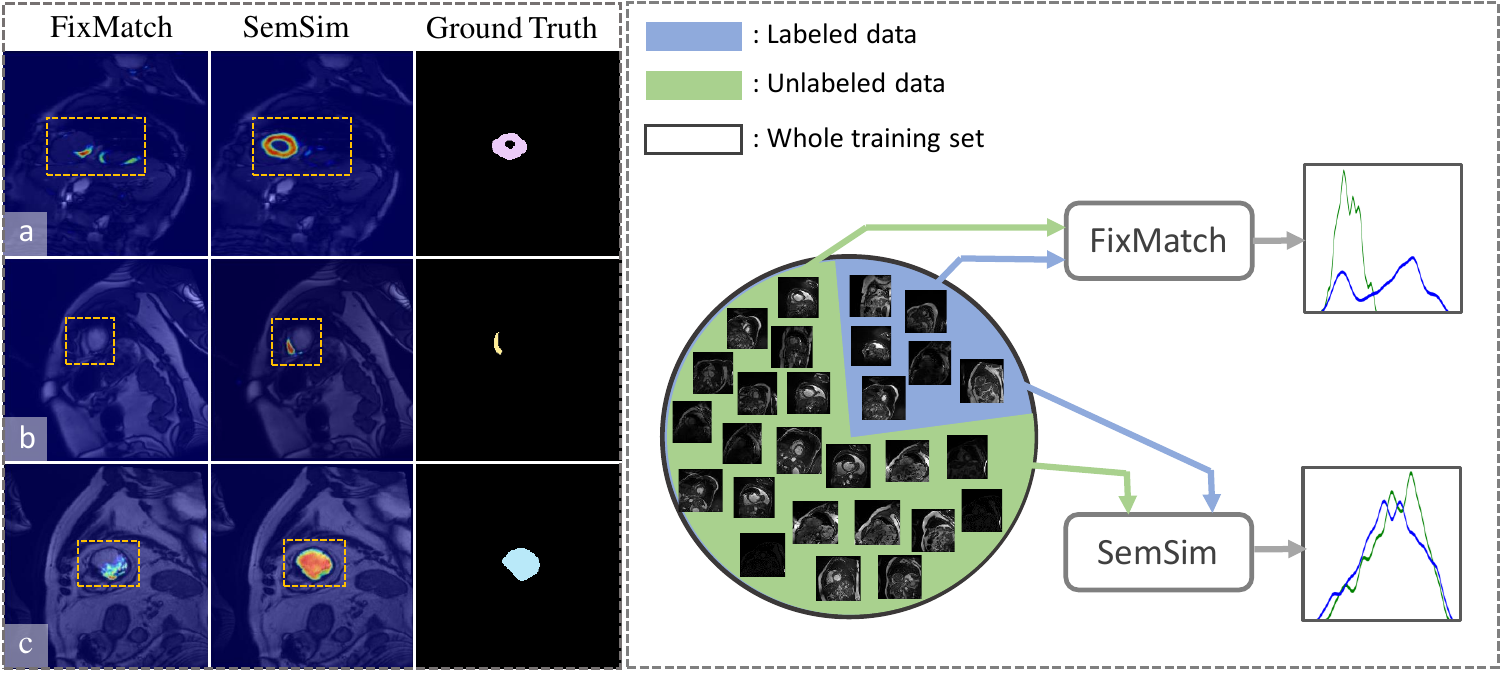}
\vspace{-10pt}
\caption{\textit{Left:} Visualization of class activation maps generated by Grad-CAM~\cite{grad} for FixMatch and SemSim. (a), (b) and (c) represent features of class Myo, RV, and LV. \textit{Right:} Kernel density estimations of voxels belonging to the Myo class in the ACDC dataset. FixMatch suffers from empirical distribution mismatch between labeled and unlabeled data, while SemSim effectively narrows the distribution gap.}
\label{fig3}
\vspace{-10pt}
\end{figure}

Unfortunately, there still exist two main issues that have been overlooked in this design: 
\textbf{(i) Intra-image problem:} 
According to the label propagation algorithm~\cite{labelpropo}, features with higher similarity should be assigned the same labels. FixMatch~\cite{sohn2020fixmatch} only focuses on pixel-wise prediction and fails to uncover this implicit pair-wise correlations in feature space within an unlabeled image, leading to discontinuities and incomplete semantic features (shown in the left column of Fig.~\ref{fig3}). 
\textbf{(ii) Cross-image problem:} 
Labeled and unlabeled data should follow the same category distribution. However, the limited volume of labeled data, along with its insufficient exploitation by conventional supervised training alone, often results in discrepancies between the class distributions learned from labeled and unlabeled data 
(shown in the right column of Fig.~\ref{fig3}). 
In light of these problems, we conclude that \textit{relying solely on existing consistency constraint is far from sufficient.}

To this end, we propose a novel SSL framework for medical image segmentation, named SemSim, from the perspective of semantic similarity. As illustrated in Fig.~\ref{fig2}~(b), instead of feeding a single strong augmented view $x^s$ into the model, we independently yield dual-stream perturbations $(x^{s1}, x^{s2})$ from unlabeled data, accompanied by two additional consistency constraints:
\textbf{(i) Intra-image semantic consistency}: 
The dependable contextual dependencies are critical for precise medical image segmentation. 
Motivated by this, we propose extracting the feature-level affinity map to refine the original pixel-wise prediction, and obtain more stable predictions, $p^{in}$ for $x^{s1}$ and $p_1^w$ for $x^w$. 
Applying the constraint between $p_1^w$ and $p^{in}$ provides explicit guidance for deep semantic features, thus enhancing the feature continuity in Fig.~\ref{fig3}. 
\textbf{(ii) Cross-image semantic consistency}: 
Medical imaging data exhibits anatomical structural similarities across different images, which creates a potential opportunity to establish a connection between labeled and unlabeled data.
Therefore, we propose to directly intervene in the unlabeled data training flow with labeled data, and leverage the cross-image semantic feature similarity to generate predictions, $p^{cr}$ for $x^{s2}$ and $p_2^w$ for $x^w$. The weak-to-strong consistency between $p^{cr}$ and $p_2^w$ in turn fully exploits reliable knowledge from labeled data thereby promoting a more compact intra-class distribution in Fig.~\ref{fig3}.

% it is clear that extracting a strong feature representation is essential for accurately calculating both intra- and cross-image feature similarities.
% 引入transformer，再强调下创新, 但是可以考虑下缩减，放在上一段的后面
Furthermore, it is evident that extracting powerful feature representations is vital for accurately calculating both intra- and cross-image feature similarities.
Previous studies have demonstrated that multi-scale features are able to capture distinctive information and address complex scale variations. Inspired by this, we develop a cross-scale feature fusion module that leverages the Transformer's~\cite{vit, swin} ability to build long-range dependencies.  
Specifically, we take into account the spatial correspondences among patches at various scales through down-sampling operations, thus generating more representative features in a lightweight manner. Benefiting from the above designs, our framework achieves superior results over state-of-the-art (SOTA) semi-supervised medical image segmentation methods, as compared in Fig.~\ref{fig1}. 
In summary, the main contributions are four-fold: 

\noindent (\textbf{\textcolor[RGB]{158, 0, 0}{i}}) 
We analyze the semantic inconsistency issues ignored by FixMatch when extended to semi-supervised medical image segmentation. Our proposed framework, SemSim, aims to strengthen the reliability of semantic features as well as  establish semantic consistencies from both intra- and cross-image perspectives.

\noindent (\textbf{\textcolor[RGB]{158, 0, 0}{ii}}) We introduce intra-image semantic consistency by imposing explicit constraints on contextual information and cross-image semantic consistency to establish cohesive class distributions.

\noindent (\textbf{\textcolor[RGB]{158, 0, 0}{iii}}) We design a lightweight spatial-aware fusion module to generate more powerful feature representations, enabling capture of more dependable correlations within the data.

\noindent (\textbf{\textcolor[RGB]{158, 0, 0}{iv}}) We conduct extensive experiments on three public medical image segmentation benchmarks, demonstrating that SemSim consistently outperforms other SSL methods. 
% It advocates for applying strong perturbations to unlabeled images and supervising the training process using predictions from weakly perturbed images.
\section{Related Works}
\label{sec:relatedworks}
\subsection{Semi-supervised Learning}
The main challenge in SSL lies in the design of effective and robust supervision signals for unlabeled data. 
Current approaches can be categorized into three major strategies: entropy minimization~\cite{EM}, pseudo-labeling~\cite{CPS, CTCT, s4cvnet}, and consistency regularization~\cite{sohn2020fixmatch, MT, ict_med, ICT, URPC, SSNet, UAMT}. Among these, the essence of consistency regularization is ensuring that a model’s predictions for the same unlabeled sample remain consistent across different perturbations. 
FixMatch~\cite{sohn2020fixmatch} stands out as a key framework based on consistency regularization. Building upon FixMatch~\cite{sohn2020fixmatch}, FlexMatch~\cite{flexmatch} integrates a curriculum learning strategy~\cite{curriculum_learning} that dynamically adjusts thresholds based on the learning state, leading to a significant reduction in training time. 
Additionally, FreeMatch~\cite{freematch} introduces a self-adaptive class fairness regularization penalty to promote diverse predictions in the early stages of training. 
Fast FixMatch~\cite{fastfixmatch} proposes utilizing fixed curriculum to regulate the unlabeled batch size, which reduces the computational load during training. These mentioned works, as variants of FixMatch~\cite{sohn2020fixmatch}, have demonstrated considerable performance improvements in semi-supervised classification tasks. Our proposed SemSim inherits from FixMatch~\cite{sohn2020fixmatch} yet addresses its limitations by focusing on semantic consistency from both intra- and cross-image perspectives, improving its adaptability to segmentation tasks.
% ICT~\cite{ICT} and ICT-Med~\cite{ict_med} propose maintaining consistency between the predictions of two interpolated unlabeled datasets and their respective interpolations. 
\subsection{Semi-supervised Medical Image Segmentation}
Semi-supervised segmentation methods~\cite{DCT, CCT, DAN, SSNet, medfct, ict_med, ICT} aim to leverage a substantial amount of unlabeled data to improve segmentation performance. One of the mainstream approaches is founded on the principle of consistency regularization. 
Mean-Teacher framework (MT)~\cite{MT} and its variants~\cite{UAMT} focus on allowing the predictions generated from either teacher
or student model as close as possible. 
BCP-Net~\cite{bcpnet} suggests learning common semantics between the labeled and unlabeled
data by a simple bidirectional copy-paste strategy. 
Another notable approach, CPS~\cite{CPS}, extends pseudo-labeling by using pseudo-labels generated by one perturbed segmentation network to supervise another. 
Recent researches highlight the exceptional performance of Transformers~\cite{swin, vit, transunet} in segmentation tasks. Inspired by this, CTCT~\cite{CTCT} integrates CNN and Transformer architectures in a cross-teaching manner, enabling the model to capture both local and global dependencies effectively. S4CVnet~\cite{s4cvnet} introduces a dual-view co-training strategy along with consistency-aware supervision. MedFCT~\cite{medfct} achieves a more powerful backbone by efficiently exploring the fusion mechanism of Transformer and CNN in the Fourier domain. CnT-B~\cite{cnt_b} designs a bi-level uncertainty estimation strategy to ensure that the learning direction is more inclined towards the reliable side. 

With the advent of large foundation models~\cite{sam, std, ho2020denoising}, 
DiffRect~\cite{diffrect} capitalizes on the power of DDPM~\cite{ho2020denoising} to learn the latent structure of the semantic labels.
CPC-SAM~\cite{CPC_SAM} leverages the prompting mechanism in SAM~\cite{sam}, which automatically generates prompts and supervisions across two decoder branches. 
To pursue simplicity and elegance, we use FixMatch~\cite{sohn2020fixmatch} as the baseline and investigate how to fully unlock its potential in the medical image segmentation task.

\section{METHODOLOGY}
\label{sec:methods}
\subsection{Prelimentaries}
FixMatch~\cite{sohn2020fixmatch} employs a weak-to-strong consistency regularization to leverage unlabeled data. 
The prediction $p^w$ of an unlabeled image $x_u$ with weak perturbation $\mathcal{A}_w$ is used to constrain the prediction $p^s$ of the same image with strong augmentation $\mathcal{A}_s$. A custom segmentation network denoted as $F$, such as UNet~\cite{unet}, is employed. The unsupervised loss $\mathcal{L}_u$ can be formulated as:
\begin{equation}
\label{eq1}
\begin{aligned}
&p^w=F(\operatorname{\mathcal{A}_w}(x_u)), \quad p^s=F(\operatorname{\mathcal{A}_s}(x_u)), \\
\mathcal{L}_u = &\frac{1}{\left | \mathcal{B}_u \right |}\sum \mathds{1} (\operatorname{max}(p^w) \ge \tau) \odot \mathcal{L}_{dice}(p^w, p^s),
\end{aligned}
\end{equation}

\noindent where $\mathcal{B}_u$ is a batch of unlabeled data, and $\tau$ is a pre-defined confidence threshold used to filter noisy pseudo labels. The supervised loss $\mathcal{L}_s$ combines the cross-entropy loss $\mathcal{L}_{ce}$ and the dice loss $\mathcal{L}_{dice}$ to minimize the difference between the prediction $p^l$ and the ground truth $y_l$, it can be formulated as:
\begin{equation}
\label{eq2}
\mathcal{L}_s = \frac{1}{2} (\mathcal{L}_{ce}(p^l, y_l) + \mathcal{L}_{dice}(p^l, y_l)).
\end{equation}
Then, the overall objective $\mathcal{L}$ combining supervised loss $\mathcal{L}_s$ and unsupervised loss $\mathcal{L}_u $ can be computed as:

\begin{equation}
\label{eq3}
\begin{aligned}
\mathcal{L} = \lambda_s \mathcal{L}_s + \lambda_u \mathcal{L}_u,
\end{aligned}
\end{equation}
where $\lambda_s$ and $\lambda_u$ are weighting coefficients that balance the supervised loss and unsupervised loss, respectively. 

\begin{figure*}[ht]
\centering
    \includegraphics[width=1.0\linewidth, height=0.29\textheight]{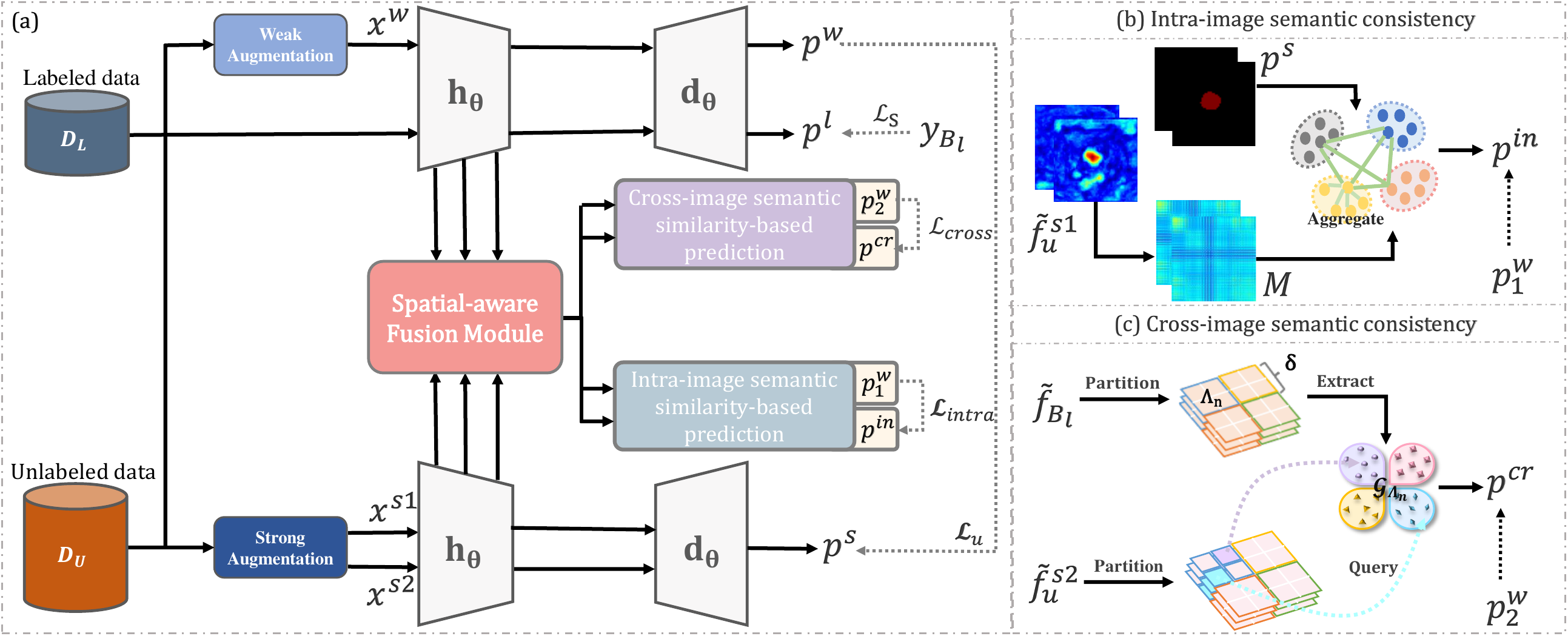} 
    \vspace{-15pt}
    \caption{(a) Overview of SemSim framework. (b) The predictions $p^{in}$, $p^{w}_1$ are based on intra-image semantic similarity. (c) The predictions $p^{cr}$, $p^{w}_2$ are based on cross-image semantic similarity. }
    \label{fig4} 
    \vspace{-10pt}
\end{figure*}

\subsection{Intra-image Semantic Consistency}
Contextual dependencies are crucial for assessing the model's training quality, where features extracted from a well-trained network should exhibit high similarity within the same class and being distinctly separable between different classes. 
However, in scenarios where labeled data is limited, relying on pixel-wise consistency constraint alone may be inadequate. 
It might fail to effectively guide a segmentation network in capturing reliable contextual dependencies from individual pixels. Thus, we opt to directly incorporate relationships extracted from the affinity map in feature-level to enhance the original pixel-wise prediction. The resulting predictions are referred as \textbf{intra-image semantic similarity-based predictions}.

As illustrated in Fig.~\ref{fig4}~(a) and (b), for an unlabeled image, the strong augmented view $x^{s1}$ is first sent into encoder $h_{\theta}$. The extracted multi-scale feature maps are then fed into the spatial-aware fusion module (described in section~\ref{D}) to generate the enhanced feature representation $\tilde{f}_{u}^{s1} \in \mathbb{R}^{D \times H \times W}$, where $D$ is the channel dimension, $H$ and $W$ denote height and width. 
After that, we reshape it to the shape of $D \times HW$ and compute the intra-image affinity map $M \in \mathbb{R}^{HW \times HW}$ as: 
\begin{equation}
\small
\label{eq4}
\begin{aligned}
M\left(k_{1}, k_{2}\right)=\operatorname{softmax}\left(\frac{\tilde{f}_{u}^{s1}\left(i_{1}, j_{1}\right)^{\top} \cdot \tilde{f}_{u}^{s1}\left(i_{2}, j_{2}\right)}{\left\|\tilde{f}_{u}^{s1}\left(i_{1}, j_{1}\right)\right\|_{2}\left\|\tilde{f}_{u}^{s1}\left(i_{2}, j_{2}\right)\right\|_{2}}\right), 
\end{aligned}
\end{equation}
where $(i_{\cdot}, j_{\cdot})$ is the coordinate of a pixel in the feature map, and $(k_{\cdot}, k_{\cdot})$ is the coordinate in the affinity map. Note that $k_1$ and $(i_1, j_1)$ denote the position of the same pixel. $M$ enables accurate delineation of the corresponding regions belonging to the same object.  

To enhance the model’s awareness of pairwise similarity, the refined prediction $p^{in} \in \mathbb{R}^{C \times HW}$, achieved by combining intra-class affinity reasoning result and the original prediction $p^{s}$, can be represented as:
\begin{equation}
\label{eq5}
\begin{aligned}
% p^{in}\left(k_{1}\right) &= p^{s}\left(k_{1}\right) + \sum_{k_{2}}^{H \times W} M\left(k_{1}, k_{2}\right) \cdot p^{s}\left(k_{2}\right), 
p^{in} = I(p^{s}) + I(p^{s}) \cdot M, 
\end{aligned}
\end{equation}
where $p^{s} \in \mathbb{R}^{C \times HW}$ is generated by segmentation decoder $d_{\theta}$. $I(\cdot)$  is a bilinear interpolation for shape matching. For the weak augmented view $x^w$, the refined prediction $p^w_1$ can also be computed with $p^w$ in the same way. 
% We define $p_1^w$ and $p^{in}$ as intra-image semantic similarity-based predictions.
Finally, we perform consistency constraint $\mathcal{L}_{intra}$ between $p_1^w$ and $p^{in}$:
\begin{equation}
\label{eq6}
\begin{aligned}
\mathcal{L}_{intra} = \frac{1}{\left | \mathcal{B}_u \right |} \sum \mathds{1} (\operatorname{max} (p^w_1) \ge \tau) \odot \mathcal{L}_{dice}(p^w_1, p^{in}).
\end{aligned}
\end{equation}
In this way, we explicitly spread the contextual dependencies into model logits output, and optimize them through a weak-to-strong consistency constraint.

\subsection{Cross-image Semantic Consistency}
The insufficient utilization of limited labeled data makes it challenging for the model to achieve consistent class representations. 
Hence, we propose explicitly guiding the learning of category-wise semantic features from both labeled and unlabeled data. Specifically, we suggest generating predictions for the unlabeled data by leveraging the trustworthy class distribution derived from the labeled data, via a dynamic feature querying mechanism. This process accelerates the transfer of dependable information from labeled data to unlabeled data, thereby narrowing the distribution gap between them. 
The obtained predictions are defined as 
\textbf{cross-image semantic similarity-based predictions}. Each step will be explained in detail in the following sections.

\subsubsection{Class Statistics of Labeled Batch $\mathcal{B}_l$} As shown in Fig.~\ref{fig4}~(c), for labeled data, once the enhanced feature map $\tilde{f}_{\mathcal{B}_l} \in \mathbb{R}^{\left | \mathcal{B}_l \right | \times D \times H \times W}$ is obtained, a typical method is to compute class prototypes directly based on masks, using this global information to represent the overall class distribution. 
However, due to the spatially heterogeneous characteristics of background, which involves various anatomical structures, this global operation will lead to an unreasonable averaging of structural information. 
Hence, given that medical images generally exhibit similar spatial layouts across different slices, we propose to partition the feature map into multiple sub-regions and calculate local class prototypes instead. As shown in Fig.~\ref{fig4}~(c), we divide $\tilde{f}_{\mathcal{B}_l}$ into a set of sub-regions $\Omega =\{\Lambda _{n}\}_{n=1}^{N}$ with the region size of $\delta \times \delta$, $N$ is the number of sub-regions. 
Within each $\Lambda _{n}$, the prototype $\mathcal{G}^c_{\Lambda_{n}}$ for each class $c$ is computed by averaging all features of the same class, which can be formulated as:
\begin{equation}
	\begin{aligned}
		\label{eq7}
		% \mathcal{G}^c_{\Lambda _{n}}=\frac{\sum_{i\in \Lambda _{n}} y_{\mathcal{B}_l}^c(i) \cdot \tilde{f}_{\mathcal{B}_l}(i)}{\sum_{i\in \Lambda _{n}} y_{\mathcal{B}_l}^c(i)}, 
        \mathcal{G}^c_{\Lambda _{n}}=\frac{\sum_{i\in \Lambda _{n}} \mathds{1}(y_{\mathcal{B}_l}(i)=c) \odot \tilde{f}_{\mathcal{B}_l}(i)}{\sum_{i\in \Lambda _{n}} \mathds{1}(y_{\mathcal{B}_l}(i)=c)}, 
	\end{aligned}
\end{equation}
where $y_{\mathcal{B}_l}$ is the class label of the entire $\mathcal{B}_l$ and $\mathcal{G}_{\Lambda _{n}} \in \mathbb{R}^{\left | \mathcal{B}_l \right | \times C \times D}$. In that way, the class prototypes obtained tends to be more accurate and reliable than global ones. 

\subsubsection{Dynamically Query Labeled Batch $\mathcal{B}_l$} As shown in Fig.~\ref{fig4}~(c), for an unlabeled image, after we obtain its enhanced feature map $\tilde{f}_u^{s2} \in \mathbb{R}^{D \times H \times W}$, we also perform the same region partition as labeled feature maps. Then, we may employ the cosine similarity between the query feature in $\tilde{f}_u^{s2}$ and the corresponding matched sub-region prototype set $\mathcal{G}_{\Lambda _{n}}$:
\begin{equation}
	\begin{aligned}
		\label{eq8}
         m_{\Lambda_n}(i, j) = \operatorname{cos} ( \tilde{f}_u^{s2}(i), \mathcal{G}_{\Lambda _{n}}(j)),
	\end{aligned}
\end{equation}
where $i \in \Lambda _{n}$ and $j \in \left | \mathcal{B}_l \right |$. Accordingly, we will obtain similarity score vector at the $i$-th position $m_{\Lambda_n}(i) \in \mathbb{R}^{\left | \mathcal{B}_l \right | \times C}$, which reflects the affinity between this feature vector and the prototypes within all batch corresponding to its sub-region. Similarly, we can apply the same operation to other sub-regions and calculate the overall similarity map $m \in \mathbb{R}^{\left | \mathcal{B}_l \right | \times C \times H \times W}$. 
The final cross-image semantic similarity based prediction $p^{cr} \in \mathbb{R}^{C \times H \times W}$ can be defined as:
\begin{equation}
	\begin{aligned}
		\label{eq9}
  p^{cr}(c) =\frac{\sum_{j=1}^{\left | \mathcal{B}_l \right |} e^{m(j,c)}}{\sum_{c} \sum_{j=1}^{\left | \mathcal{B}_l \right |} e^{m(j,c)}}. 
	\end{aligned}
\end{equation}
% batch中每个图的不确定性
% $w_{s}^{v}=e^{-\mathcal{D}_{s}^{v}}$
% 相当于每个通道都有个不确定性，然后如何得到整体的不确定性

\subsubsection{Uncertainty Estimation of $p^{cr}$} 
In the initial phase of training, feature learning can exhibit considerable uncertainty. To effectively assess the stability of the training process and the consistency of predictions, we evaluate the uncertainty of the obtained cross-image semantic similarity-based predictions. Specially, we first compute the average prediction by querying a batch of labeled data:
\begin{equation}
	\begin{aligned}
		\label{eq10}
\bar{m}=\frac{1}{\left | \mathcal{B}_l \right |} \sum_{j=1}^{\left | \mathcal{B}_l \right |} m(j). 
\end{aligned}
\end{equation}
The uncertainty of the $j$-th prediction by inferring the $j$-th labeled image can be calculated by the KL-divergence (KL):
\begin{equation}
	\begin{aligned}
		\label{eq11}
U(j)=\operatorname{KL} \left[m(j) \| \bar{m}\right]=\sum_{c=0}^{C-1}m(j,c) \log \frac{m(j,c)}{\bar{m}(c)}. 
\end{aligned}
\end{equation}
Such pixel-level uncertainty reflects the approximate variance that assesses the difference between the batch-wise and the averaged one, where a larger value indicates lower similarity. The average uncertainty $\bar{U} \in \mathbb{R}^{H \times W}$ across batch of labeled images for the prediction $p^{cr}$ can be defined as:
\begin{equation}
	\begin{aligned}
		\label{eq12}
\bar{U} = \frac{1}{\left | \mathcal{B}_l \right |} \sum_{j=1}^{\left | \mathcal{B}_l \right |}e^{{-r \times U(j)}}, 
\end{aligned}
\end{equation}
where $r$ is a hyperparameter. The corresponding weak-to-strong consistency loss will be enforced on this prediction by incorporating the uncertainty computed before:
\begin{equation}
\label{eq13}
\begin{aligned}
\mathcal{L}_{cross} &= \frac{1}{\left | \mathcal{B}_u \right |}\sum \mathds{1} (\operatorname{max}(p^w_2) \ge \tau) \odot  \mathcal{L}_{ce} (p^w_2, p^{cr}) \odot \bar{U},
\end{aligned}
\end{equation}
where $p_2^w$ is also obtained by calculating the cross-image semantic similarity. It should be noted that when the differences between labeled and unlabeled batches are significant, \textit{e.g.} the local prototype of a specific class is absent in a certain sub-region, we will prioritize using the class prototype of the entire image as a substitute. 

\subsection{Spatial-aware Fusion Module}
\label{D}
\begin{figure}[ht]
\centering
    \includegraphics[width=1.0\linewidth, height=0.16\textheight]{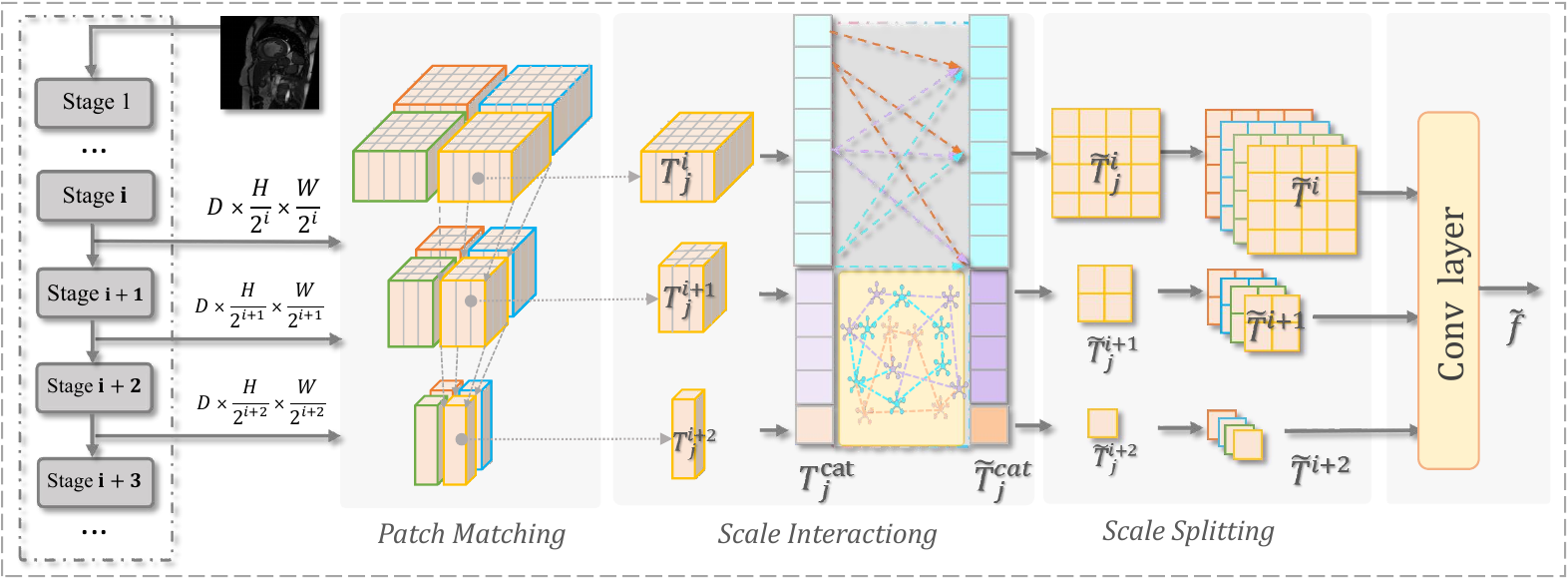} 
    \vspace{-15pt}
    \caption{Overview of spatial-aware fusion module. $H$, $W$ represent the height and width of the feature map.}
    \label{fig5} 
    \vspace{-14pt}
\end{figure}

Multi-scale cues have been shown to be highly effective in addressing complex scale variations, allowing for the capture of both global and local contextual information of the target. The key challenge lies in how to efficiently integrate this information to construct a robust feature representation, which is essential for the reliability of our proposed semantic similarity-based predictors. To tackle this, we develop a spatial-aware transformer-based block that facilitates interaction across different scales in an efficient manner. Fig.~\ref{fig5} illustrates the specific fusion process from the $i$-th to the $(i+2)$-th scale.

Rather than computing dependencies within a lengthy sequence directly concatenated from tokens at multiple scales (obtained by flattening the entire feature map of each scale), we focus on the long-range cross-scale dependency of the object itself and the nearby objects. First, we harness the spatial relationships among patches in different scales, called \textbf{\textit{Patch Matching}}. As shown in Fig.~\ref{fig5}, we denote $T_{j}^i$ as the $j$-th patch at the $i$-th scale, and the corresponding down-sampled patches are $T_{j}^{(i+1)}$ and $T_{j}^{(i+2)}$ sharing the bounding box with the same color (\textit{e.g.} yellow). We first regularize feature maps into the same channel dimension, then concatenate the corresponding inter-scale patches, which can be formulated as:
\begin{equation}
\label{eq14}
\left[\operatorname{flatten}\left(T_j^i\right), \operatorname{flatten}\left(T_j^{(i+1)}\right), \operatorname{flatten}\left(T_j^{(i+2)}\right)\right] \rightarrow T_j^{\text {cat }},
\end{equation}
where $\operatorname{flatten}(\cdot)$ rearranges the patch size into 1D sequence and $\left[\cdot\right]$ represents the concatenation operation. In this way, we focus on the most related patches, and thus maintain the spatial correspondence as well as reduce the redundancy. 
The second step \textbf{\textit{Scale Interacting}} that aims to capture dependencies among patches can obtained as follows:
\begin{equation}
\label{eq15}
\begin{aligned}
& \hat{T}_j^{cat}=\operatorname{MSA}\left(\operatorname{LN}\left(T_j^{c a t}\right)\right)+T_j^{c a t}, \\
&\tilde{T}_j^{cat}=\operatorname{MLP}\left(\operatorname{LN}\left(\hat{T}_j^{c a t}\right)\right)+\hat{T}_j^{cat},
\end{aligned}
\end{equation}
\noindent where $\operatorname{LN}(\cdot)$ denotes the Layer Normalization~\cite{ba2016layer}, $\operatorname{MSA}(\cdot)$ represents Multi-head Self-Attention operation and $\operatorname{MLP}(\cdot)$ denotes a two-layer linear projection. 
After that, we will utilize \textbf{\textit{Scale Splitting}} to convert the enhanced sequence back to patches based on the concatenation order:
\begin{equation}
\label{eq16}
\operatorname{Reshape}\left(\operatorname{Split}\left(\tilde{T}_j^{\text {cat }}\right)\right) \rightarrow \tilde{T}_j^i, \tilde{T}_j^{(i+1)}, \tilde{T}_j^{(i+2)}, 
\end{equation}
where $\operatorname{Split}(\cdot)$ is the inverse operation of previous concatenation operation. Finally, we perform interpolation ($I(\cdot)$) on the enhanced feature maps to uniform the spatial dimension, and then employ convolution layers to fuse these interpolated feature maps and obtain the final feature representation $\tilde{f}$. This process is described as follows:
\begin{equation}
\label{eq17}
\tilde{f}=\operatorname{RELU}(\operatorname{BN}(\operatorname{Conv}([I(\tilde{T}^i), \tilde{T}^{(i+1)}, I(\tilde{T}^{(i+2)})]))),
\end{equation}
where $\operatorname{BN}(\cdot)$ represents BatchNormalization layer~\cite{Batch_normalization} and $\operatorname{RELU(\cdot)}$~\cite{relu} is an activation function.

\vspace{3pt}
\noindent \textbf{Complexity Analysis}. 
If we directly concatenate the multi-scale feature maps and apply an original transformer block, the computational complexity is $O(\frac{H^2W^2D}{16^{i}})$, where the sequence length is $\left( \frac{HW}{2^{2i}} + \frac{HW}{2^{2i+2}} + \frac{HW}{2^{2i+4}} \right)$.
Our proposed spatial-aware fusion module first splits the feature map at each scale into $S \times S$ windows, with each window having a size of $\frac{H}{2^i \times S} \times \frac{W}{2^i \times S}$ in the $i$-th stage. By leveraging cross-scale spatial-aware correspondences, we only need to perform self-attention $S \times S$ times, with a sequence length given by $\frac{1}{S^2} \left( \frac{HW}{2^{2i}} + \frac{HW}{2^{2i+2}} + \frac{HW}{2^{2i+4}} \right)$ for each operation.
Thus, the complexity of each operation is approximately $O\left( \frac{H^2W^2D}{16^{i}S^4} \right)$, and the total computational load becomes $O\left( \frac{H^2W^2D}{16^{i}S^2} \right)$.

\begin{algorithm*}[t!]
\small
\caption{Pseudocode of SemSim.}
\label{alg1}
    \KwIn{Batch of unlabeled data $\mathcal{B}_{u} =\left \{ x_u \right \}$ / Batch of labeled data $\mathcal{B}_{l} =\left \{ x_l \right \}$. 
    Encoder $h_{\theta}$ and decoder $d_{\theta}$ of network $F$.} 
    \KwOut{Fully trained model $F$ for inference.}
    \For{$x_u\in \mathcal{B}_{u}$}
    {
        $x^w = \mathcal{A}_w(x_u)$ \quad $x^{s1}$, $x^{s2}=\mathcal{A}_s(x_u),\mathcal{A}_s(x_u)$;
        $\quad \qquad \qquad \qquad$ 
        \textcolor[RGB]{23, 118, 26}{\# Get weak and two different strong augmented views of $x_u$}\;

        $\tilde{f}_u^{s1} = \mathrm{SFM} (h_\theta(x^{s1}))$, $\tilde{f}_u^{s2} = \mathrm{SFM} (h_\theta(x^{s2}))$, $\tilde{f}_{\mathcal{B}_l} = \mathrm{SFM} (h_{\theta} ({\mathcal{B}_l}))$, $\tilde{f}_u^{w} = \mathrm{SFM} (h_\theta(x^{w}))$;
        $\quad$
        \textcolor[RGB]{23, 118, 26}{\# Extract the enhanced features}\;

        $p^w = F(x^w)$ \quad \quad $p^{s} = F(x^{s1})$;
        $\qquad \qquad \qquad \quad \qquad \qquad$ 
        \textcolor[RGB]{23, 118, 26}{\# Obtain weak and strong augmented predictions through decoder}\;

        Obtain the intra-image semantic similarity-based predictions $p_1^w$ / $p^{in}$ using $\tilde{f}_u^{w}$ /$\tilde{f}_u^{s1}$ based on 
        Eq.~(\textcolor[RGB]{255, 0, 0}{4})-(\textcolor[RGB]{255, 0, 0}{5})\;

        Obtain the cross-image semantic similarity-based predictions $p_2^w$ / $p^{cr}$ using ($\tilde{f}_u^{w}$, $\tilde{f}_{\mathcal{B}_l}$) / ($\tilde{f}_u^{s2}$,$\tilde{f}_{\mathcal{B}_l}$) based on Eq.~(\textcolor[RGB]{255, 0, 0}{7})-(\textcolor[RGB]{255, 0, 0}{9})\;

        Calculate the intra-image semantic consistency constraint $\mathcal{L}_{intra}$ using $p_1^w$ and $p^{in}$ based on 
        Eq.~(\textcolor[RGB]{255, 0, 0}{6})\;
        
        Calculate the cross-image semantic consistency constraint $\mathcal{L}_{cross}$ using $p_2^w$ and $p^{cr}$ based on 
        Eq.~(\textcolor[RGB]{255, 0, 0}{10})-(\textcolor[RGB]{255, 0, 0}{13})\;
        
        Calculate the total unsupervised loss $\tilde{\mathcal{L}}_{u}$ based on 
        Eq.~(\textcolor[RGB]{255, 0, 0}{18})\;
    }
    \For{$x_l\in \mathcal{B}_l$}
    {
        Calculate the supervised loss $\mathcal{L}_s$ based on Eq.~(\textcolor[RGB]{255, 0, 0}{2})\;
    }
    
    Train $F$ by averaging $\mathcal{L}_s$ and $\tilde{\mathcal{L}}_{u}$\;

    \textbf{return} $F$
\end{algorithm*}

\label{algorithm}

\subsection{Training and Inference}
\subsubsection{Loss Function} In summary, based on the original unsupervised loss (Eq.~\ref{eq1}) in FixMatch~\cite{sohn2020fixmatch}, our SemSim further incorporates two additional consistency losses from semantic similarity perspective: $\mathcal{L}_{intra}$ (Eq.~\ref{eq6}) and $\mathcal{L}_{cross}$ (Eq.~\ref{eq13}).
Therefore, the unsupervised loss can be rephrased as:
\begin{equation}
\begin{split}
		\label{eq18} 
        \tilde{\mathcal{L}}_{u} &= \lambda \mathcal{L}_{u}  + \lambda_{intra}\mathcal{L}_{intra} + \lambda_{cross}\mathcal{L}_{cross}
\end{split}
\end{equation}

where the loss weights $\lambda$, $\lambda_{intra}$, $\lambda_{cross}$ are set to 0.5, 0.25, 0.25, respectively. Consequently, the overall objective can be described in the following, where the supervised loss $\mathcal{L}_s$ stays consistent with FixMatch~\cite{sohn2020fixmatch}:
\begin{equation}
	\begin{aligned}
           \begin{split}
		\label{eq19} 
        \mathcal{L} &= \frac{1}{2} ({\mathcal{L}_s + \tilde{\mathcal{L}}_{u}}).
        \end{split}
\end{aligned} 
\end{equation}

\subsubsection{The Algorithm of SemSim} The pseudocode of our SemSim framework is presented in Algorithm~\ref{algorithm}. This framework takes an equal number of labeled and unlabeled images as input, and aims to obtain the fully-trained $F$ for inference. During inference process, assuming the test input image is $x_t$, we just need to pass it through the encoder $h_\theta$ and decoder $d_\theta$ of $F$ to obtain the final prediction.

\section{Experiments}
\subsection{Datasets and Evaluation Metrics}
In our experiments, we use three commonly used datasets to verify the effectiveness of our method, including ACDC dataset~\cite{bernard2018deep}, ISIC dataset~\cite{codella2019skin} and PROMISE12 dataset~\cite{promise}.

\textbf{(i) ACDC dataset} is a cardiac MRI dataset that contains 200 annotated short-axis cardiac MRI images from 100 patients with three organs. Following~\cite{CTCT}, we split the dataset into 70 patients for training, 10 patients for validating and 20 patients for testing, respectively. For semi-supervised training, we evaluate different methods with 1\% (1 case), 5\% (3 case) and 10\% (7 case) labeled separately. 

\textbf{(ii) ISIC dataset} is a skin lesion segmentation dataset with 2 classes. We use 1838 images for training and the rest 756 images for validation. Under a semi-supervised setting, 3\% (55 images), 10\% (181 images) of training data are provided with labels, while the rest of training images are unlabeled. 

\textbf{(iii) PROMISE12 dataset} is available for the MICCAI 2012 prostate segmentation challenge. T2-weighted MRIs of 50 patients with various conditions are acquired at different locations. The dataset is divided into 35 training cases, 5 validation cases, and 10 test cases. All 3D scans are converted into 2D slices. For the semi-supervised setting, $3\%$ and $7\%$ are regarded as labeled, while the rest remain unlabeled.

Three commonly used metrics are employed to evaluate the segmentation results: (1) Dice Coefficient (DSC), (2) 95\% Hausdorff Distance (95HD), and (3) Average Symmetric Surface Distance (ASSD). The DSC measures the overlap between the prediction and the ground truth, while the 95HD calculates the distances between the surfaces of the prediction and the ground truth. Following previous approaches~\cite{scpnet}, DSC and 95HD are used to evaluate performance on the ACDC and ISIC datasets, whereas DSC and ASSD are utilized as evaluation metrics for the PROMISE12 dataset.

\subsection{Implementation Details}
Following~\cite{CTCT}, UNet~\cite{unet} is adopted as the backbone. We resize all the slices to $224 \times 224$. Our augmentation strategy is consistent with~\cite{CTCT}, where color transformation and CutMix~\cite{yun2019cutmix} are set as strong augmentations, and random flipping is utilized as weak augmentation. During training, the batch size is set to $16$, consisting of an equal number of unlabeled and labeled cases. We optimize our network with the SGD optimizer~\cite{sgd}, where the weight decay is set to $1 \times 10^{-4}$ and the momentum is set to 0.9. 
All experiments are trained for $300$ epochs. Moreover, the learning rate is initially set to $0.01$. To broaden the perturbation space, we further introduce feature level augmentation by adopting a channel dropout of 50\% probability. The channel dimension $D$ is set to $128$ in our setting. $r$ and $\tau$ are set to 1000 and 0.95. 

\begin{table*}[t]
\small
\renewcommand\arraystretch{1.0}
\centering
\caption{Comparison of SemSim with other SSL methods on ACDC dataset under different ratios of labeled data. -- represents that the value is too unstable and low to write, which remains below 0.3 after retraining multiple times. }
\scalebox{1.0}{
\setlength{\tabcolsep}{3.7mm}{
\begin{tabular}{lcccccc|c}
\toprule
\multirow{2}{*}{Method} & \multicolumn{2}{c}{1\% labeled}    & \multicolumn{2}{c}{5\% labeled}    & \multicolumn{2}{c|}{10\% labeled}   & \multirow{2}{*}{Params (M)} \\ \cline{2-7} 
& DSC $\uparrow$& 95HD (mm) $\downarrow$ & DSC $\uparrow$  & 95HD (mm) $\downarrow$& DSC $\uparrow$  & 95HD (mm) $\downarrow$  \\ \midrule
Only sup        & $0.390$          & $54.7$              & 0.560          & 39.8              & 0.797          & 9.8               & 2.60 \\ \midrule
MT~\cite{MT}            & -              & -                 & 0.566          & 34.5              & 0.810          & 14.4              & 1.81 \\ 
EM~\cite{EM}      & -              & -                 & 0.602          & 24.1              & 0.791          & 14.5          & 1.81 \\ 
UA-MT~\cite{UAMT}                   & -              & -                 & 0.610          & 25.8              & 0.815          & 14.4              & 1.81 \\ 
DCT~\cite{DCT}                     & -              & -              & 0.582          & 26.4              & 0.804          & 13.8              & 1.81 \\ 
CCT~\cite{CCT}                     & -              & -                 & 0.586          & 27.9              & 0.816          & 13.1              & 3.71 \\ 
CPS~\cite{CPS}                     & -              & -                 & 0.603          & 25.5              & 0.833          & 11.0              & 3.62 \\ 
ICT~\cite{ICT}                     & -              & -                 & 0.581          & 22.8              & 0.811          & 11.4              & 1.81 \\ 
DAN~\cite{DAN}                     & -              & -                 & 0.528          & 32.6              & 0.795          & 14.6              & 1.81 \\ 
URPC~\cite{URPC}                    & -              & -                 & 0.567          & 31.4              & 0.829          & 10.6              & 1.83 \\ 
CTCT~\cite{CTCT}                    & -              & -                 & 0.704          & 12.4              & 0.864          & 8.6               & 29.11 \\ 
SSNet~\cite{SSNet}                   & -              & -                 & 0.705          & 17.4              & 0.853          & 10.6              & 1.81 \\ 
ICT-Med~\cite{ict_med}                 & -              & -                 & 0.563          & 22.6              & 0.837          & 13.1              & 1.81 \\ 
S4CVNet~\cite{s4cvnet}                 & 0.534          & 37.2              & 0.731          & 5.1               & 0.873          & 3.9               & 55.31 \\ 
FixMatch~\cite{sohn2020fixmatch}      & 0.722          & 22.8              & 0.824          & 4.5               & 0.879                    & 3.1               & 1.81 \\ 
MedFCT~\cite{medfct}    & 0.528    & 24.1        & 0.726       & 10.5  &0.886 & 4.3 & 31.18 \\ 
CnT-B~\cite{cnt_b}       & 0.632   & 19.3    & 0.755 & 10.6  & 0.880 & 5.5 & 29.11 \\ 
BCP-Net~\cite{bcpnet}    & 0.649    & 18.0    & 0.879      &  {\color[HTML]{0000FF}2.1}     & 0.889      & 4.0  & 1.81 \\ 
DiffRect~\cite{diffrect}     & 0.720      & {\color[HTML]{0000FF}5.8}   & 0.836  & 7.6    & {\color[HTML]{0000FF}0.891}     & 3.9 & 19.37 \\ 
CPC-SAM~\cite{CPC_SAM}     & {\color[HTML]{0000FF}0.856}      & 9.2   & {\color[HTML]{0000FF}0.880}  & 5.8    & 0.890     & {\color[HTML]{0000FF}3.1} & 93.75 \\  
\midrule
SemSim$^{-}$    & 0.861     &  4.3      & 0.882     & 2.4    &  0.893                & 2.7 & 1.81 \\ 
SemSim                    & {\color[HTML]{FF0000}0.872}          & {\color[HTML]{FF0000}1.8}               & {\color[HTML]{FF0000}0.888}          & {\color[HTML]{FF0000}1.9}               & {\color[HTML]{FF0000}0.896}          & {\color[HTML]{FF0000}2.3} & 2.60 \\ 
\bottomrule
\end{tabular}}}
\label{tab1}
\vspace{-12pt}
\end{table*}

\begin{table}[t]
\renewcommand\arraystretch{1.1}
\centering
\small
\caption{Comparison of SemSim with other SSL methods on ISIC dataset under different ratios of labeled data.}
\scalebox{0.95}{
\setlength{\tabcolsep}{1.5mm}{
\begin{tabular}{lcccc}
\toprule
\multirow{2}{*}{Method} & \multicolumn{2}{c}{3\% labeled}    & \multicolumn{2}{c}{10\% labeled}   \\ \cline{2-5} 
                        & DSC $\uparrow$ & 95HD (mm) $\downarrow$ & DSC $\uparrow$ & 95HD (mm) $\downarrow$ \\ \hline
Only sup                & 0.663          & 28.4              & 0.691          & 26.3              \\ \midrule
MT~\cite{MT}                        & 0.728          & 37.4              & 0.734          & 34.0              \\
EM~\cite{EM}                        & 0.723          & 36.3              & 0.727          & 39.3              \\
UA-MT~\cite{UAMT}                   & 0.730          & 38.6              & 0.734          & 33.2              \\
DCT~\cite{DCT}                     & 0.729          & 40.6              & 0.760          & 35.7              \\
CCT~\cite{CCT}                     & 0.677          & 42.2              & 0.723          & 31.7              \\
CPS~\cite{CPS}                     & 0.686          & 44.4              & 0.743          & 35.7              \\
ICT~\cite{ICT}                     & 0.732          & 37.2              & 0.753          & 34.6              \\
DAN~\cite{DAN}                     & 0.695          & 39.5              & 0.724          & 30.4              \\
URPC~\cite{URPC}                    & 0.703          & 39.3              & 0.758          & 32.8              \\
CTCT~\cite{CTCT}                    & 0.713          & 43.2              & 0.760          & 37.3              \\
SSNet~\cite{SSNet}                   & 0.728          & 40.8              & 0.758          & 32.8              \\
ICT-Med~\cite{ict_med}                 & 0.714          & 39.2              & 0.749          & 33.1              \\
S4CVNet~\cite{s4cvnet}                 & 0.752          & 37.2              & 0.774          & 31.8              \\
MedFCT~\cite{medfct}        & 0.742     & 39.0         & 0.780      & 29.6                              \\
FixMatch~\cite{sohn2020fixmatch}       & 0.756    & 36.7        & 0.783       & {\color[HTML]{0000FF}15.5}             \\
CnT-B~\cite{cnt_b}       & 0.762     & 29.6    & 0.792    &  25.0                            \\
BCP-Net~\cite{bcpnet}    & {\color[HTML]{0000FF}0.769}    & {\color[HTML]{0000FF}24.5}    & {\color[HTML]{0000FF}0.794}    & 18.7 \\
SemSim           & {\color[HTML]{FF0000}0.774}          & {\color[HTML]{FF0000}18.9}              & {\color[HTML]{FF0000}0.802}          & {\color[HTML]{FF0000}13.3}              \\ 
\bottomrule
\end{tabular}}}
\label{tab2}
\vspace{-16pt}
\end{table}

\begin{table}[t]
\renewcommand\arraystretch{1.1}
\centering
\small
\caption{Comparison of our SemSim with other SSL methods on 3\% and 7\% labeled data of PROMISE12 dataset.}
\scalebox{1.0}{
\setlength{\tabcolsep}{1.2mm}{
\begin{tabular}{lcccc}
\toprule
\multirow{2}{*}{Method} & \multicolumn{2}{c}{3\% labeled}    & \multicolumn{2}{c}{7\% labeled}   \\ \cline{2-5} 
                        & DSC $\uparrow$ & ASSD (mm) $\downarrow$ & DSC $\uparrow$ & ASSD (mm)$\downarrow$ \\ \hline
Only sup      & 0.552    & 12.8   & 0.587   & 8.4              \\ \midrule
MT~\cite{MT}        & 0.405   & 3.1   & 0.714  & 7.6                  \\ 
UA-MT~\cite{UAMT}   & 0.552   & 8.6   & 0.657  & 2.4                  \\ 
CCT~\cite{CCT}      & 0.467   & 2.1   & 0.714  & 16.7                 \\  
URPC~\cite{URPC}    & 0.619  & 2.9   & 0.632  & 4.3                  \\ 
SSNet~\cite{SSNet}  & 0.574   & 6.3   & 0.623  & 4.4                  \\ 
SCP-Net~\cite{scpnet}    & 0.551   & 19.5  & 0.771 & 3.5     \\
S4CVnet~\cite{s4cvnet}   & 0.628   & {\color[HTML]{0000FF} 1.9}   & 0.672 & 3.1                \\
FixMatch~\cite{sohn2020fixmatch}   & 0.463  & 4.2   & 0.716  & 6.7  \\
CnT-B~\cite{cnt_b}       & 0.685   & 5.4   & 0.763  & 2.7                     \\
MedFCT~\cite{medfct}     & {\color[HTML]{0000FF} 0.725}      & 2.2   &0.760 &1.8             \\
BCP-Net~\cite{bcpnet}    & 0.706   & 8.8   & {\color[HTML]{0000FF} 0.772}      & {\color[HTML]{0000FF} 1.4}                \\
SemSim & {\color[HTML]{FF0000}0.758}     & {\color[HTML]{FF0000}1.6}   & {\color[HTML]{FF0000}0.784}   &  {\color[HTML]{FF0000}1.3}    \\ 
\bottomrule
\end{tabular}}}
\label{tab3}
\vspace{-16pt}
\end{table}

\subsection{Comparison with State-of-the-Art}
\subsubsection{Quantitative Comparison}
We compare our proposed SemSim with 20 state-of-the-art semi-supervised learning methods, including (1) CNN-based methods: MT~\cite{MT}, UA-MT~\cite{UAMT}, EM~\cite{EM}, DCT~\cite{DCT}, CCT~\cite{CCT}, CPS~\cite{CPS}, ICT~\cite{ICT}, DAN~\cite{DAN}, URPC~\cite{URPC}, SSNet~\cite{SSNet}, ICT-Med~\cite{ict_med}, FixMatch~\cite{sohn2020fixmatch}, SCPNet~\cite{scpnet}, BCP-Net~\cite{bcpnet}; (2) hybrid methods (CNN and Transformer): CTCT~\cite{CTCT}, S4CVnet~\cite{s4cvnet}, MedFCT~\cite{medfct}, CnT-B~\cite{cnt_b}; (3) generative method: DiffRect~\cite{diffrect}; and (4) SAM-based method: CPC-SAM~\cite{CPC_SAM}. Following~\cite{CTCT}, we conduct all experiments under the same settings on the public ACDC, ISIC, and PROMISE12 datasets. "Only sup" in Tables~\ref{tab1},~\ref{tab2}, and~\ref{tab3} refers to training with labeled data only.

\textbf{Results on ACDC dataset:} The quantitative comparison results on the ACDC dataset are shown in Table~\ref{tab1}. Specifically, with 5\% and 10\% labeled data, SemSim demonstrates remarkable improvements compared to the previous state-of-the-art method CPC-SAM~\cite{CPC_SAM} (DSC: +0.8\%, +0.6\%). Notably, even with only 1\% labeled data, SemSim still substantially outperforms existing approaches in both the regional measure DSC (+1.6\%) and the boundary-aware measure 95HD (-7.4mm). This highlights that our method offers a significant advantage in scenarios with minimal labeled data, where most of the other methods perform poorly (with DSC below 0.3). Besides, our SemSim contains only 2.60 million parameters, substantially fewer than several high-performing methods such as DiffRect~\cite{diffrect} and CPC-SAM~\cite{CPC_SAM}. 

We further eliminate the interference of spatial-aware fusion module (SFM) and replace it with a simpler convolution (SemSim$^{-}$). It can be observed that even without this module, our model equipped with intra- and cross-image semantic consistencies already outperforms existing methods in all partitions. Due to the lightweight design of SFM, which consists of only 0.79 million parameters, SemSim achieves significant improvement, particularly with 1\% labeled data.
\begin{figure*}[t]
	\centering
	\includegraphics[width=1.0\linewidth, height=0.27\textheight]{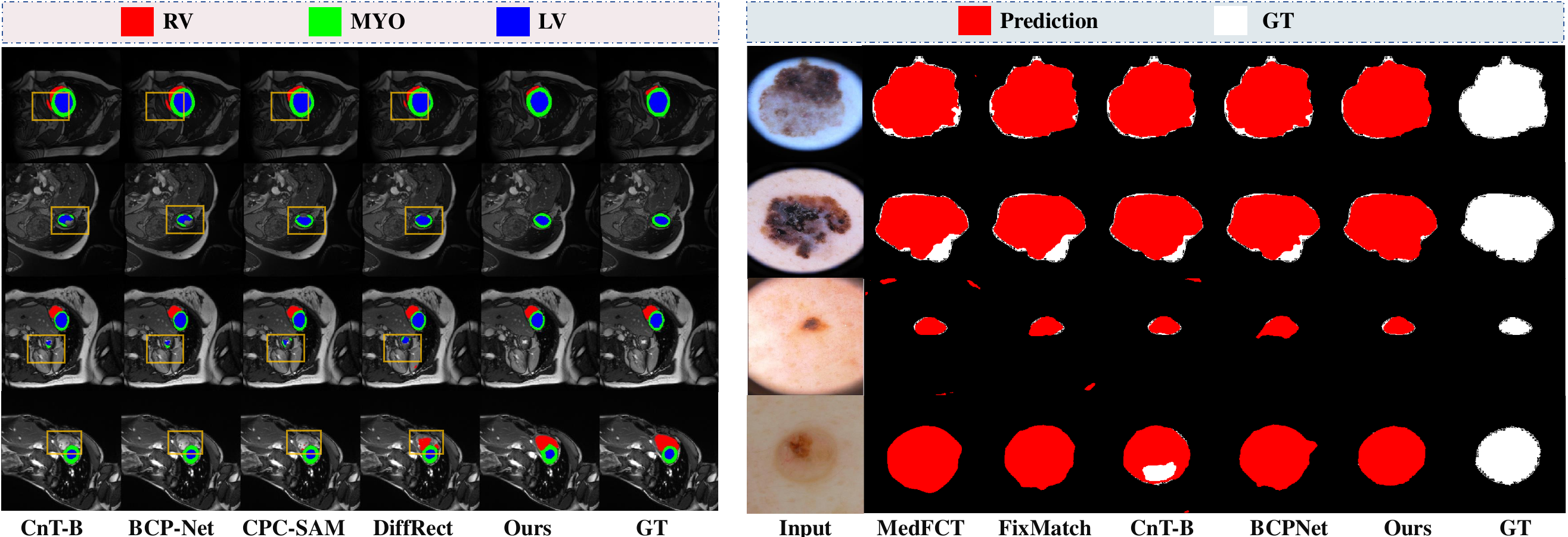}
        \vspace{-15pt}
	\caption{Comparison of segmentation results on the ACDC (left) and ISIC (right) datasets with 10\% labeled data. } 
	\label{fig6}
\vspace{-10pt}
\end{figure*} 

\begin{figure}[t]
	\centering
	\includegraphics[width=0.98\linewidth, height=0.19\textheight]{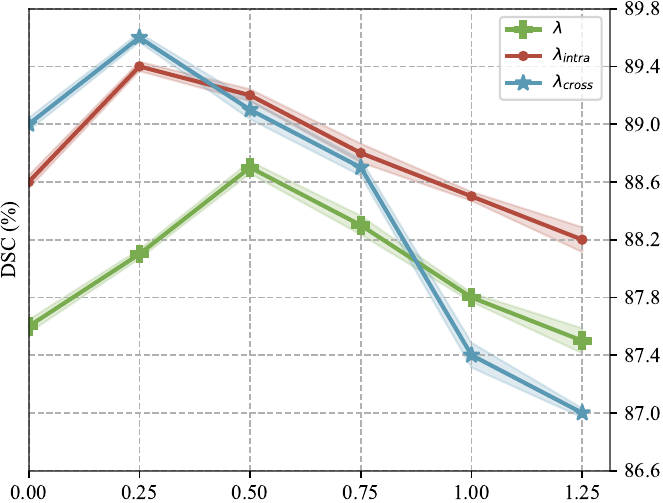}
        \vspace{-5pt}
	\caption{Performance of SemSim \text{w.r.t.} $\lambda$, $\lambda_{intra}$ and $\lambda_{cross}$ on the ACDC dataset.} 
	\label{fig7}
\vspace{-10pt}
\end{figure} 

\begin{figure}[t]
	\centering
	\includegraphics[width=1.0\linewidth, height=0.17\textheight]{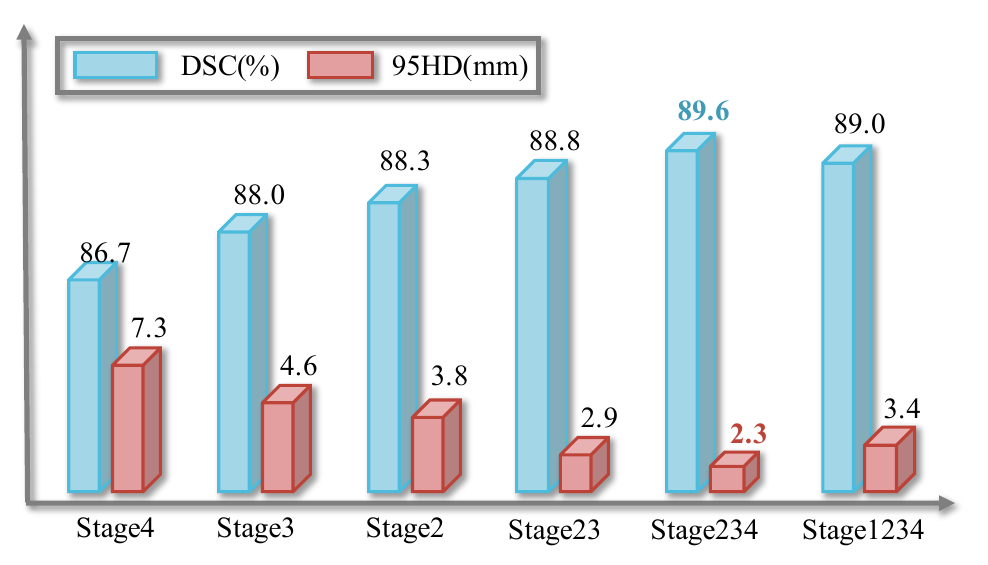}
        \vspace{-20pt}
	\caption{Ablation of spatial-aware fusion module on different stages on the ACDC dataset with 10\% labeled.}
	\label{fig8}
\vspace{-15pt}
\end{figure}

\begin{table}[t]
\renewcommand\arraystretch{1.0}
\small
\centering
\caption{Ablation study of different consistency constraints on 1\% labeled data of ACDC dataset.}
\scalebox{1.0}{
\setlength{\tabcolsep}{2.9mm}{
\begin{tabular}{c|c|c|c|c|c}
\toprule
\multicolumn{1}{l|}{\#} & \multicolumn{1}{l|}{$\mathcal{L}_{intra}$} & \multicolumn{1}{l|}{$\mathcal{L}_{cross}$}  & \multicolumn{1}{l|}{$\mathcal{L}_{u}$}  & \multicolumn{1}{l|}{DSC $\uparrow$} & \multicolumn{1}{l}{95HD (mm) $\downarrow$}  \\ 
\midrule
1   & $\times$  & $\times$       & $\checkmark$    & 0.835   & 7.6       \\ 
2   & $\checkmark$  & $\times$       & $\times$    & 0.844   & 5.0       \\ 
3   & $\times$      & $\checkmark$   & $\times$    & 0.824   & 5.5          \\ 
4   & $\checkmark$  & $\checkmark$    & $\times$    & 0.861  & 4.3       \\  
5  & $\checkmark$   &  $\times$      & $\checkmark$  & 0.868    & 2.9       \\ 
6  & $\times$   &  $\checkmark$      & $\checkmark$  & 0.862    &  4.1      \\ 
7  & $\checkmark$   &  $\checkmark$      & $\checkmark$  &  {\color[HTML]{FF0000}0.872}   & {\color[HTML]{FF0000}1.8}       \\ 
\bottomrule
\end{tabular}}}
\vspace{-10pt}
\label{tab5}
\end{table}

\begin{table}[t]
\renewcommand\arraystretch{1.2}
\centering
\caption{Effect of the number of sub-regions $N$ on ACDC, ISIC and PROMISE12 datasets.}
\scalebox{1.0}{
\setlength{\tabcolsep}{0.6mm}{
\begin{tabular}{c|cc|cc|cc}
\toprule
\multicolumn{1}{c|}{} & \multicolumn{2}{c|}{ACDC (10\%)} & \multicolumn{2}{c|}{ISIC(10\%)} & \multicolumn{2}{c}{PROMISE12 (7\%)} \\ \cline{2-7} 
\multicolumn{1}{c|}{\multirow{-2}{*}{N}}
&\multicolumn{1}{c}{DSC $\uparrow$ }
&\multicolumn{1}{c|}{95HD (mm) $\downarrow$}        
&\multicolumn{1}{c}{DSC $\uparrow$}         
&\multicolumn{1}{c|}{95HD (mm) $\downarrow$}        
&\multicolumn{1}{c}{DSC $\uparrow$} 
&\multicolumn{1}{c}{ASSD (mm) $\downarrow$}         \\
\midrule
1    & 0.888     & 3.6   & 0.775      & 18.1     & 0.770   &  3.4            \\
2   &  0.891     & 2.9   & 0.795     &  16.5     & 0.778   &  2.1           \\
4   &  {\color[HTML]{FF0000}0.896}     & {\color[HTML]{FF0000}2.3}  & {\color[HTML]{FF0000}0.802}   & {\color[HTML]{FF0000}13.3}             & {\color[HTML]{FF0000}0.784}   & {\color[HTML]{FF0000}1.3}                   \\
8   &  0.893     & 2.4   & 0.794   & 15.3       & 0.780         &  2.9            \\
14  &  0.887     & 4.3   &  0.789   & 17.1      & 0.775         &  4.5            \\
\bottomrule
\end{tabular}}}
\label{tab4}
\vspace{-15pt}
\end{table}

\textbf{Results on ISIC dataset:} Table~\ref{tab2} presents the quantitative results of different methods on the ISIC dataset. Our framework, compared to training with only labeled data (Only sup), effectively leverages unlabeled data to enhance segmentation accuracy. Consistent with the results in Table~\ref{tab1}, SemSim outperforms the previous best method, BCP-Net~\cite{bcpnet}, with improvements in DSC (+0.5\% and +0.8\%) and reductions in 95HD (-5.6mm and -5.4mm) under the 3\% and 10\% labeled data settings, respectively, establishing a new state-of-the-art. These results on both datasets substantiate the robustness and effectiveness of our method.

\textbf{Results on PROMISE12 dataset:} Table~\ref{tab3} presents the performance comparison between SemSim and other SOTA methods on the PROMISE12 dataset, using labeled data percentages of 3\% and 7\%. Compared with training with only labeled data, SemSim can achieve +20.6\% and 19.7\% improvements with 3\% and 7\% labeled. 
Although recent approaches such as MedFCT\cite{medfct} and BCP-Net\cite{bcpnet} have shown promising results, they fall short in effectively delineating clear boundaries and establishing distinct category relationships. 
Our SemSim, which integrates newly proposed intra- and cross-image semantic consistency constraints with a powerful feature fusion module, achieving the highest performance 75.8\% and 78.4\% under 3\% and 7\% labeled settings on this dataset.

\subsubsection{Visual Comparisons}
Fig.~\ref{fig6} illustrates the qualitative results of different methods on the ACDC dataset~\cite{bernard2018deep} with 10\% labeled and ISIC dataset~\cite{codella2019skin} with 10\% labeled, including FixMatch~\cite{sohn2020fixmatch}, MedFCT~\cite{medfct}, CnT-B~\cite{cnt_b}, BCPNet~\cite{bcpnet}, CPC-SAM~\cite{CPC_SAM}, DiffRect~\cite{diffrect}, ours and GroundTruth. 
Benefiting from the intra- and cross-image semantic consistencies design, SemSim is capable of generating satisfactory segmentation results closer to the ground truth with only 10\% labeled, ensuring both the integrity and accuracy of the segmented targets. 
The phenomenon
can be explained by its ability to enhance feature continuity and strengthen the distinctions between categories. 

\subsection{Hyper-Parameters}
\subsubsection{Impact of Loss Function Weights}
$\lambda$, $\lambda_{intra}$ and $\lambda_{cross}$ are three coefficients that balance the overall unsupervised loss. We explore the sensitivity of the remaining weight by fixing two of them. As shown in Fig.~\ref{fig7}, we first fix $\lambda_{intra}=0.25$ and $\lambda_{cross}=0.25$, it is evident that the accuracy peaks at $\lambda=0.5$, highlighting that the conventional pixel-wise consistency plays a more significant role in model performance. Next, with $\lambda_{cross} = 0.25$ and $\lambda = 0.5$, the optimal performance of 89.4\% is achieved at $\lambda_{intra} = 0.25$. Finally, we examine the appropriate value for $\lambda_{cross}$ while fixing $\lambda = 0.5$ and $\lambda_{intra} = 0.25$. 
As the weight $\lambda_{cross}$ increases, we observe a sharp decline in accuracy primarily due to the overfitting of unlabeled data to the limited labeled data. Thus, the optimal combination of loss weights $\lambda$, $\lambda_{intra}$, $\lambda_{cross}$ is 0.5, 0.25, 0.25.
\begin{table}[t]
\renewcommand\arraystretch{1.2}
\centering
\caption{Ablation study of strong perturbed streams on ACDC and ISIC datasets with 10\% labeled data.}
\scalebox{1.0}{
\setlength{\tabcolsep}{1.25mm}{
\begin{tabular}{ccccccc}
\toprule
\multirow{2}{*}{$p^{in}$} & \multirow{2}{*}{$p^{cr}$} & \multirow{2}{*}{$p^{s}$} & \multicolumn{2}{c}{ACDC (10\% labeled)}              & \multicolumn{2}{c}{ISIC (10\% labeled)} \\ \cline{4-7} 
&             &         & \multicolumn{1}{c}{DSC $\uparrow$} & \multicolumn{1}{c}{95HD (mm) $\downarrow$} & DSC $\uparrow$  & 95HD (mm) $\downarrow$ \\
\midrule
S1    & S1        & S1     & 0.890        & 2.0                                & 0.790     & 20.0       \\
S1                 & S1        & S2     & 0.884        & 2.7                                                & 0.786  & 19.8        \\
S1                 & S2        & S1     & {\color[HTML]{FF0000}0.896}   & 2.3            
                   & {\color[HTML]{FF0000}0.802}  & {\color[HTML]{FF0000}13.3}         \\
S1                 & S2        & S2          & 0.877     & 1.9         
                   & 0.779     & 15.8         \\
S1                 & S2        & S3           & 0.887     & 1.5       
                   & 0.755     & 31.1          \\
$\times$             & $\times$    & S1/S2      & 0.868      & 2.1
                   & 0.780     & 19.9              \\
$\times$             & $\times$   & S1/S2/S3   & 0.889     &  {\color[HTML]{FF0000}1.4}              
                   & 0.795     & 16.5
\\ \bottomrule
\end{tabular}}}
\label{tab6}
\vspace{-5pt}
\end{table}

\subsubsection{Spatial-aware Fusion Module on Different Stages} 
Additionally, we examine the impact of the spatial-aware fusion module on performance across various scales, ranging from stage 1 (size of $224 \times 224$) to stage 4 (size of $28 \times 28$). As illustrated in Fig.~\ref{fig8}, leveraging features from shallower stages leads to an improvement from 86.7\% to 88.3\%, as it encompasses both local details and adequate semantic information. The method achieves its highest performance (DSC: 89.6\%, 95HD: 2.3mm) when applying multi-scale fusion across consecutive stages (stage 2, 3 and 4). However, incorporating fusion at stage 1 can compromise intrinsic characteristics and increase computational overhead. Therefore, we implement multi-scale fusion at stages 2, 3 and 4, resulting in optimal performance while maintaining a reasonable computational complexity.
% 80.06   16.13

\subsection{Ablation Study}
\subsubsection{Effect of Different Consistency Losses}
We conduct experiments to explore the effectiveness of different consistency constraints. 
In Table~\ref{tab5}, we begin by assessing the influence of intra-image semantic consistency constraint $\mathcal{L}_{intra}$ (\#2) and cross-image semantic consistency constraint $\mathcal{L}_{cross}$ (\#3).
It indicates that with $\mathcal{L}_{intra}$ only (\#2), our performance has already surpassed that of $\mathcal{L}_{u}$ (\#1). 
Then, relying solely on the correlation with limited labeled data (\#3), our method can still achieve competitive performance (DSC: 0.824, 95HD: 5.5mm), which validates the effectiveness of $\mathcal{L}_{cross}$. 
By combining these two types of consistency constraints (\#4), our method improves by +1.7\% and +3.7\% compared to using (\#2 and \#3) separately, demonstrating the complementarity of $\mathcal{L}{intra}$ and $\mathcal{L}_{cross}$.
Moreover, we further individually integrate $\mathcal{L}_{cross}$ and $\mathcal{L}_{intra}$ with $\mathcal{L}_{u}$ (\#5 and \#6), performances have both been improved compared with using only $\mathcal{L}_{u}$ (\#1). 
Finally, SemSim achieves the best performance by combining these three consistency losses.
\begin{figure}[t]
	\centering
	\includegraphics[width=1.0\linewidth, height=0.15\textheight]{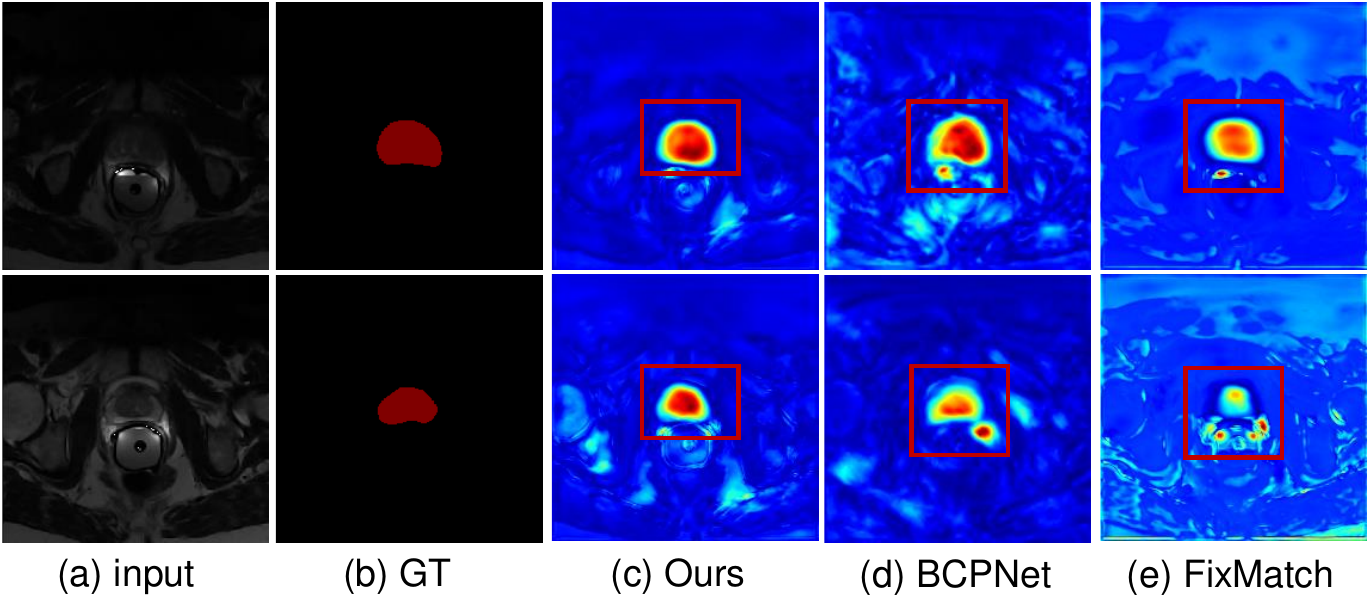}
     \vspace{-20pt}
	\caption{Feature visualizations of different methods on PROMISE12 dataset.} 
	\label{fig9}
\vspace{-8pt}
\end{figure} 

\begin{figure}[t]
	\centering
	\includegraphics[width=1.0\linewidth, height=0.16\textheight]{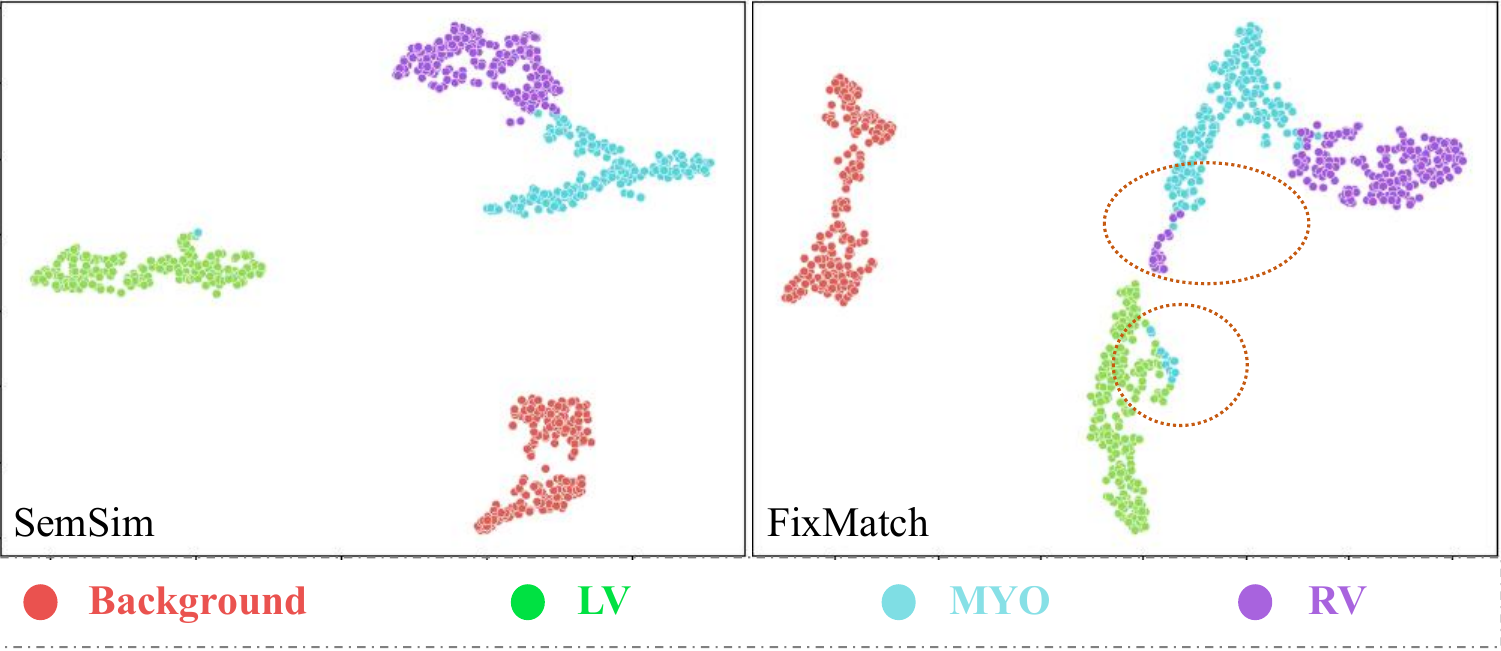}
     \vspace{-15pt}
	\caption{T-SNE visualization of deep feature representations extracted from SemSim and FixMatch on ACDC dataset. } 
	\label{fig10}
     \vspace{-10pt}
\end{figure} 

\begin{figure*}[t]
	\centering
	\includegraphics[width=1.0\linewidth, height=0.10\textheight]{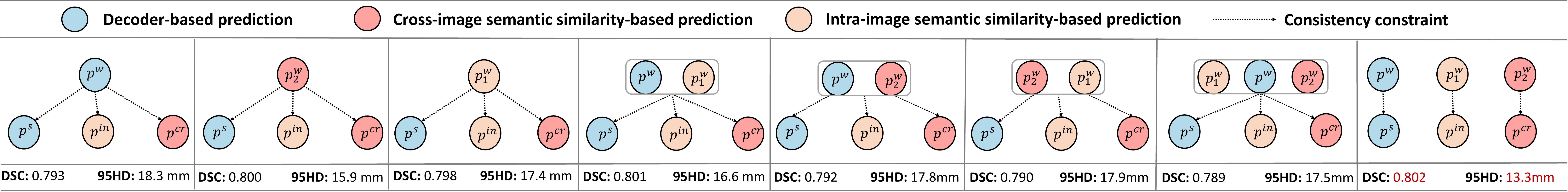}
\vspace{-15pt}
\caption{Ablation of consistency constraint framework on the ISIC dataset with 10\% labeled.}
\label{fig11}
\vspace{-15pt}
\end{figure*} 

\subsubsection{Effect of the Number of Sub-regions}
We explore the impact of the number of sun-regions $N$ on these three datasets, it can be observed from the Table~\ref{tab4} that when the number is set to $4$, the best performances are achieved across all three datasets. As the number increases, there is an escalation in computational demand, while the accuracy experiences a corresponding decline, which could be attributed to the small size of the grid regions, making it challenging for unlabeled images to capturing accurate and representative class-specific features from such limited areas. 

\subsubsection{Ablation of Strong Perturbation Streams} 
We also evaluate the impact of providing different strong augmented views as input to the three types of predictions. As shown in Table~\ref{tab6}, S1, S2, and S3 represent views subjected to varying degrees of strong data augmentation. The results indicate that using the same view for both $p^s$ and $p^{in}$ yields the highest accuracy on the ACDC and ISIC datasets. Besides, when predictions are made solely through the decoder, the triple-view setup (S1/S2/S3) outperforms the dual-view setup (S1/S2) due to the introduction of diverse perturbations. However, they are still inferior to our proposed comprehensive predictions integrating both intra- and cross-semantic similarity-based predictions, which fully leverage image-level perturbations.

\subsubsection{Ablation of Consistency Constraint Framework} 
As illustrated in Fig.~\ref{fig11}, we can obtain three types of predictions: through decoder ($p^w$, $p^{s}$), intra-image semantic similarity-based predictions ($p^w_1$, $p^{in}$), cross-image semantic similarity-based predictions ($p^w_2$, $p^{cr}$). 
Then, we analyze the combinations among these predictions to achieve three types of weak-to-strong consistencies $\mathcal{L}_{intra}$, $\mathcal{L}_{cross}$ and $\mathcal{L}_u$. 
Our results indicate that averaging $p_1^w$ and $p^w$ as the ensemble prediction for $x^w$ achieves superior performance, with DSC: 0.801 and 95HD: 16.6mm. Notably, for each prediction type, the corresponding weak view prediction acts as a pseudo label to guide the strongly augmented predictions, yielding the highest accuracy of 0.802 on the ISIC dataset.

\subsection{Interpretation of our SemSim}
We visualize the class activation maps of various methods in Fig.~\ref{fig9}. Compared to BCPNet\cite{bcpnet} and FixMatch~\cite{sohn2020fixmatch}, SemSim inherits the advantages of intra-image semantic consistency that considers the relationships between pixels, reducing pixel-level misclassification and resulting in more complete object segmentation. As illustrated in Fig.~\ref{fig10}, the pixel embeddings learned by SemSim become more compact and well-separated.  It demonstrates that the designed cross-image semantic consistency effectively extracts distinctive features from the labeled data, even with a limited amount of data available. 

\section{Conclusion}
In this paper, we propose a novel semi-supervised medical image segmentation framework, termed SemSim, which addresses the intra- and cross-image semantic inconsistency challenges faced by FixMatch. We thoroughly investigate two key consistency mechanisms based on semantic similarity: one emphasizes contextual dependencies between internal features within an image to refine final predictions, thereby achieving more continuous segmentation. The other extracts semantic relationships between labeled and unlabeled data, enabling the model to learn more accurate and consistent class distributions even with limited labeled data. Further, an efficient spatial-aware fusion module is introduced to assist the above consistencies by generating powerful feature representations. SemSim was extensively evaluated on three public datasets and consistently outperformed other SSL approaches.

\section*{References}
\bibliographystyle{IEEEtran}
\bibliography{refs}
\end{document}